\documentclass{article}
\usepackage[preprint]{log_2024}			

\usepackage{booktabs}						
\usepackage{multirow}						
\usepackage{amsfonts}						
\usepackage{graphicx}						
\usepackage{blindtext}
\usepackage{xcolor}

\newcommand{\method}{PyG-SSL}

\definecolor{LightGray}{gray}{0.9}

\usepackage[numbers,compress,sort]{natbib}	

\title{PyG-SSL: A Graph Self-Supervised Learning Toolkit}

\author[L. Zheng et al.]{
Lecheng Zheng\thanks{Equal Contribution.},~ Baoyu Jing\footnotemark[1],~ Zihao Li\footnotemark[1], Zhichen Zeng, Tianxin Wei, Mengting Ai,
Xinrui He, \\
\normalsize{\textbf{Lihui Liu, Dongqi Fu, Jiaxuan You, Hanghang Tong, Jingrui He}} \\
University of Illinois Urbana-Champaign, Wayne State University, Meta AI \\
\footnotesize{\texttt{\{lecheng4, baoyuj2, zihaoli5, zhichenz, twei10, mai10, xhe33\}@illinois.edu}} \\
\footnotesize{\texttt{hw6926@wayne.edu}}, 
\footnotesize{\texttt{dongqifu@meta.com}}, \footnotesize{\texttt{\{jiaxuan, htong, jingrui\}@illinois.edu}}
}

\begin{document}

\maketitle

\begin{abstract}
Graph Self-Supervised Learning (SSL) has emerged as a pivotal area of research in recent years. By engaging in pretext tasks to learn the intricate topological structures and properties of graphs using unlabeled data, these graph SSL models achieve enhanced performance, improved generalization, and heightened robustness. Despite the remarkable achievements of these graph SSL methods, their current implementation poses significant challenges for beginners and practitioners due to the complex nature of graph structures, inconsistent evaluation metrics, and concerns regarding reproducibility hinder further progress in this field. Recognizing the growing interest within the research community, there is an urgent need for a comprehensive, beginner-friendly, and accessible toolkit consisting of the most representative graph SSL algorithms. To address these challenges, we present a Graph SSL toolkit named \method, which is built upon PyTorch and is compatible with various deep learning and scientific computing backends. Within the toolkit, we offer a unified framework encompassing dataset loading, hyper-parameter configuration, model training, and comprehensive performance evaluation for diverse downstream tasks. Moreover, we provide beginner-friendly tutorials and the best hyper-parameters of each graph SSL algorithm on different graph datasets, facilitating the reproduction of results. The GitHub repository of the library is \url{https://github.com/iDEA-iSAIL-Lab-UIUC/pyg-ssl}.
\end{abstract}

\section{Introduction and Motivation}

Graph Self-Supervised Learning (SSL) has emerged as a pivotal area of research in recent years. By engaging in pretext tasks to learn the intricate topological structures and properties of graphs using unlabeled data~\citep{DBLP:journals/tkde/LiuJPZZXY23}, these graph SSL models achieve enhanced performance, improved generalization, and heightened robustness~\citep{DBLP:journals/corr/abs-1809-10341, DBLP:conf/icml/HassaniA20, DBLP:conf/kdd/QiuCDZYDWT20, DBLP:conf/iclr/HuLGZLPL20, DBLP:conf/cikm/FuXLTH20, DBLP:conf/icml/YouCSW21, DBLP:conf/cikm/ZhouZF0H22, DBLP:conf/TMLR/ZhengFMH24,
jing2021hdmi,wang2024mastering,he2023robust}. Leveraging the representations learned through SSL, graph downstream tasks like node classification, similarity search, and graph classification benefit significantly~\citep{DBLP:conf/ijcai/JinZL00P21, DBLP:conf/aaai/ParkK0Y20, DBLP:conf/nips/YouCSCWS20, DBLP:conf/iclr/SunHV020,feng2022adversarial,feng2024ariel,li2022graph,he2024co}.
For instance, Deep Graph InfoMax (DGI), proposed by
Velickovic et al.~\cite{DBLP:conf/iclr/VelickovicCCRLB18}
, aims to maximize the global-local mutual information and capture the high-level summaries of graph. Mo et al.~\citep{DBLP:conf/aaai/MoPXS022} propose a multi-loss approach to capitalize on the structural and neighbor information, thus enriching interclass variation.
GraphCL~\cite{DBLP:conf/nips/YouCSCWS20} designs various graph augmentation strategies to augment the raw graph and then maximize the mutual information between two augmented graphs via InfoNCE-style contrastive loss.
Hou et al.~\citep{DBLP:conf/kdd/HouLCDYW022} introduce a masked graph autoencoder with generative self-supervised graph learning, specifically tailored to address issues inherent in graph autoencoder. 

Despite the remarkable achievements of these graph SSL methods, their current implementations pose significant challenges for beginners and practitioners due to the complex nature of graph structures and inconsistent evaluation metrics.  Additionally, concerns regarding reproducibility continue to hinder progress in this field. Recognizing the growing interest within the research community, there is an urgent need for a comprehensive, beginner-friendly, accessible toolkit that includes the most representative graph SSL algorithms.

To address these challenges, we introduce a Graph SSL toolkit, which includes a diverse set of state-of-the-art graph self-supervised learning methods tailored for various downstream tasks, such as node classification and graph classification. The toolkit offers a unified framework for dataset loading, hyper-parameter configuration, model training, and comprehensive performance evaluation. Built on PyTorch, it is compatible with various deep learning and scientific computing backends, including Numpy, PyTorch Geometric, and DGL.

\textbf{Main contributions}. (1) We present the PyTorch Graph Self-Supervised Learning Toolkit (\method), an open-source library for graph self-supervised learning, complete with public releases, documentation, code examples, and continuous integration. (2) We provide a unified framework within PyG-SSL for dataset loading, hyper-parameter configuration, model training, and performance evaluation across various downstream tasks. (3) We offer beginner-friendly tutorials and optimal hyper-parameters for different graph SSL algorithms on various datasets to facilitate result reproduction. (4) We evaluate the performance of numerous graph self-supervised learning methods, providing insights for selecting appropriate methods for specific tasks.
\section{Comparison to Related Libraries}
Table~\ref{tab:comparisons} compares the key features of existing graph SSL toolkits, where most of them are open-source code delivered with technical papers. DIG-SSL~\citep{DBLP:conf/nips/0001XLW21} is a unified and highly customizable framework for implementing graph SSL methods.
PyGCL \citep{DBLP:journals/pami/XieXZWJ23} is a graph contrastive learning library. 
These two packages bear the following limitations: (i) They only support a few SSL algorithms, i.e., 4 in DIG-SSL and 7 in PyGCL, and all these SSL algorithms are designed only for homogeneous graphs. 
Different from them, our toolkit supports many more SSL algorithms for various types of graphs, including homogeneous graphs, heterogeneous graphs, and molecular graphs; (ii) PyGCL does not provide user-friendly tutorials, which poses great challenges to beginners; (iii) Graph augmentation is a key feature in graph self-supervised learning~\citep{DBLP:conf/icml/YouCSW21}, which is missing in DIG-SSL; (iv) Our toolkit addresses the crucial reproducibility by offering the best hyper-parameters for different datasets in configuration files, which is absent in both DIG-SSL and PyGCL.


\begin{table*}[t]
    \centering
    \caption{Comparison among different graph SSL toolkits.}
    \scalebox{0.95}{
    \begin{tabular}{cccccc}
    \toprule 
    \multirow{2}{*}{}          & \# of SSL    &  Augmentation    & Various Evaluation  & Various Graph &  Beginner-friendly \\
                 & Algorithms & Module & Metrics &   Types & Tutorials  \\ 
                 \midrule 
    Ours         & 10    & \textcolor{ForestGreen}{$\checkmark$}    & \textcolor{ForestGreen}{$\checkmark$}    & \textcolor{ForestGreen}{$\checkmark$}       & \textcolor{ForestGreen}{$\checkmark$}    \\ 
    DIG-SSL      & 4     & \textcolor{red}{$\times$}    & \textcolor{ForestGreen}{$\checkmark$}    & \textcolor{red}{$\times$}       & \textcolor{ForestGreen}{$\checkmark$}    \\ 
    PyGCL        & 7     & \textcolor{ForestGreen}{$\checkmark$}    & \textcolor{red}{$\times$}    & \textcolor{red}{$\times$}       & \textcolor{red}{$\times$}    \\ 
    \bottomrule
    \end{tabular}
    }
    \label{tab:comparisons}
\end{table*}

\section{Related Works}
Graphs are ubiquitous and are widely used for ranking \cite{page1999pagerank, DBLP:conf/nips/HeTMS12, DBLP:conf/www/LiFH23, li2024apex}
, social network analysis \cite{DBLP:conf/kdd/FuZH20, DBLP:conf/www/Fu0MCBH23, DBLP:journals/corr/abs-2410-12126,zeng2023parrot,yan2024pacer,xuslog,zeng2023generative, DBLP:conf/icde/Zheng0TXZH24, DBLP:journals/fdata/ZhouZXH19, DBLP:conf/kdd/ZhouZ0H20}, recommendation \cite{DBLP:journals/corr/abs-2411-01410, DBLP:conf/kdd/WeiH22, wei2021model, wei2020fast, DBLP:conf/www/HeLKH24, wei2024towards, DBLP:conf/www/YooZKQZ0WXCT24,jing2024sterling}, question answering \cite{liu2021kompare,binet,prefnet,cornnet}, spatial-temporal modeling \cite{jing2024causality,jing2021network,jing2022retrieval,yan2021dynamic} and text summarization \cite{jing2021multiplex} etc.
Graph Representation Learning (GRL) aims to transform graph-structured data into low-dimensional representations, while preserving the graph's structural and semantic information. DGI~\cite{DBLP:conf/iclr/VelickovicFHLBH19} maximizes mutual information between subgraph representations and global graph summaries, enabling effective reuse for downstream node-level tasks. BGRL~\cite{DBLP:conf/iclr/ThakoorTAADMVV22} learns node representations by encoding two augmented graph views using an online encoder and a target encoder, where the online encoder predicts the target encoder's representations. InfoGraph~\cite{DBLP:conf/iclr/SunHV020} maximizes the mutual information between the graph-level representation and the
representations of substructures of different scales. GraphMAE~\cite{DBLP:conf/kdd/HouLCDYW022} focuses on feature reconstruction with a masking strategy and scaled cosine error to enable robust training via generative self-supervised graph pretraining.

Contrastive learning~\cite{DBLP:conf/kdd/ZhengXZH22, weiconnecting, DBLP:conf/sdm/ZhengZH23, DBLP:conf/kdd/ZhengJLTH24, DBLP:conf/icml/ChenK0H20, DBLP:journals/corr/abs-2402-02357, DBLP:journals/corr/abs-1807-03748,
liu2024contrastivelearningrefineembeddings,jing2024automated} has garnered significant attention from researchers for its exceptional performance in modeling unlabeled data. Many recent works on graph contrastive learning~\cite{DBLP:conf/nips/YouCSCWS20, DBLP:conf/icml/YouCSW21, wei2022augmentations, DBLP:conf/www/ZhengCHC24, li2024can,jing2021hdmi,jing2022x,jing2022coin,yan2024reconciling} demonstrate the effectiveness of applying contrastive learning to learn the high quality representation in unsupervised, semi-supervised and transfer learning settings. For instance, GraphCL~\cite{DBLP:conf/nips/YouCSCWS20}, JOAO~\cite{DBLP:conf/icml/YouCSW21} and GCA~\cite{DBLP:conf/www/0001XYLWW21} study the impact of graph augmentations when minimizing graph contrastive learning loss.
Graph Contrastive Coding (GCC)~\cite{DBLP:conf/kdd/QiuCDZYDWT20} uses contrastive learning for subgraph instance discrimination within and across networks to capture universal topological properties across multiple networks. MVGRL~\cite{DBLP:conf/icml/HassaniA20} demonstrates that learning representation by contrasting node and graph across different views outperforms approaches that contrast graph-graph or multi-scale encodings.  
SUGRL~\cite{DBLP:conf/aaai/MoPXS022} enhances inter-class variation by integrating structural and neighbor information while minimizing intra-class variation using an upper bound loss, achieving efficiency by eliminating data augmentation and discriminators in contrastive learning. This paper aims to offer a unified framework for dataset loading, hyper-parameter configuration, model training, and comprehensive performance evaluation of these methods.

\begin{figure}
    \centering
    \includegraphics[width=0.85\linewidth]{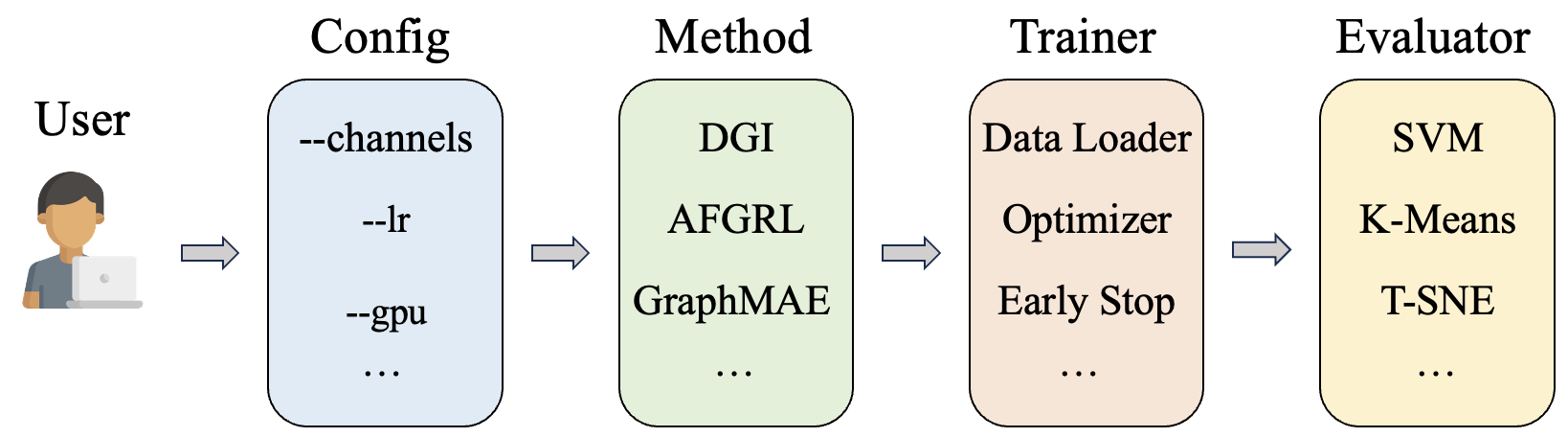}
    \caption{Overview of PyG-SSL.}
    \label{fig:overview}
\end{figure}

\section{Design of \method}
Our \method\ is designed as a user-friendly Python library for general audiences of beginners and practitioners who are interested in graph SSL methods and the applications of these methods in downstream tasks. 
As illustrated in Figure \ref{fig:overview}, our framework can be divided into four components:
\begin{itemize}
    \item \textbf{Configuration}: Users can specify detailed configurations for GNNs, SSL methods, etc.
    \item \textbf{Method}: A variety of SSL methods, e.g., DGI, can be called to pre-train GNNs.
    \item \textbf{Trainer}: A trainer that automatically trains the graph SSL method based on the provided configurations.
    \item \textbf{Evaluator}: Multiple downstream evaluators, such as support vector machine (SVM), are readily available for users to assess the performance of pre-trained GNNs.
\end{itemize}

In addition, our library provides various augmentations, similarity functions, and losses used in the SSL literature for users to build their own SSL methods. 
Under the MIT License, our toolkit relies solely on open-source libraries.


We will now provide a detailed explanation of each component in the framework.

\textbf{Configuration.}
\method\ provides an easy-to-use configuration function to load the hyper-parameters for data loader, model, optimizer, and classifier. The configurations of the data loader include the dataset name, root path to load the dataset, and batch size. The configurations of the optimizer contain the learning rate, optimizer name, maximal training epoch, patience epoch, and weight decay. Model configurations mainly consist of default hyperparameters for training. The configurations of the classifier include the hyper-parameters for model evaluation in the downstream task. To facilitate the reproduction of results, we provide beginner-friendly tutorials and the optimal hyper-parameters of each graph SSL algorithm on different graph datasets. 

\textbf{Method.}
The method module in \method\ aims to integrate the major functions of different graph self-supervised techniques. Specifically, the graph self-supervised techniques include augmentation methods for graph contrastive learning (e.g., data augmentation~\cite{DBLP:conf/nips/YouCSCWS20, DBLP:conf/icml/YouCSW21} or embedding augmentation~\cite{yu2022graph}), graph encoders to learn hidden representation (e.g., GIN~\cite{DBLP:conf/iclr/XuHLJ19}, GCN~\cite{DBLP:conf/iclr/KipfW17}, GAT~\cite{DBLP:conf/iclr/VelickovicCCRLB18}) and various loss functions for model training (e.g., InfoNCE-style loss~\cite{DBLP:conf/nips/YouCSCWS20}, masked reconstruction loss~\cite{DBLP:conf/kdd/HouLCDYW022}, etc.). The implemented graph self-supervised methods include DGI~\cite{DBLP:journals/corr/abs-1809-10341}, GraphCL~\cite{DBLP:conf/nips/YouCSCWS20}, AFGRL~\cite{lee2022augmentation}, MVGRL~\cite{DBLP:conf/icml/HassaniA20}, GCA~\cite{DBLP:conf/www/0001XYLWW21}, BGRL~\cite{DBLP:conf/iclr/ThakoorTAADMVV22}, SUGRL~\cite{DBLP:conf/aaai/MoPXS022}, ReGCL~\cite{DBLP:conf/aaai/JiHSPFLL24}, InfoGraph~\cite{DBLP:conf/iclr/SunHV020}, GraphMAE~\cite{DBLP:conf/kdd/HouLCDYW022}, etc.

\textbf{Trainer.} The trainer module takes the data loader and the graph self-supervised learning as input and trains the model until convergence. Additionally, we implement early stopping criteria to reduce the computational cost by allowing a limited number of additional epochs if training loss no longer decreases.

\textbf{Evaluator.} \method\ integrates task-specific evaluation methods, evaluation metrics and utilities for various purposes. Specifically, we provide traditional classifiers for supervised evaluation, including logistic regression, SVM, random forests, etc; the unsupervised evaluation includes k-mean clustering and similarity search. Our library also provides some utility functions tailored to the research papers we have implemented, such as embedding visualization, allowing users to easily visualize embeddings learned by different graph self-supervised methods.


\subsection{Documentation and Code Example} 
\begin{figure}
    \centering
    \includegraphics[width=0.85\linewidth]{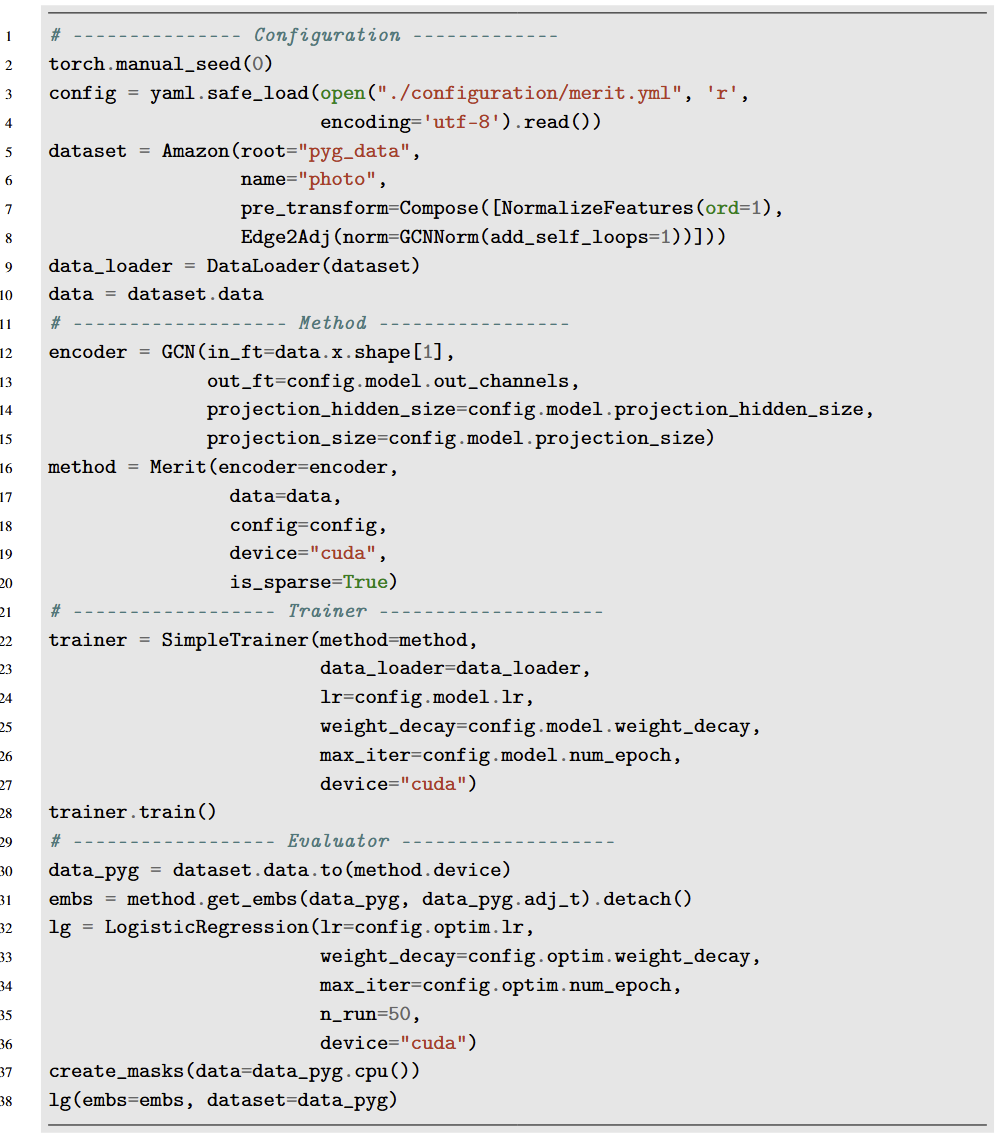}
    \caption{Example of Implementation.}
    \label{fig:code_example}
\end{figure}

We release our software with publicly available documentation at \url{https://pygssl-tutorial.readthedocs.io/en/latest/}.
Our documentation covers the four major components of \method\ and graph self-supervised methods in various research papers. The documentation also includes an in-depth tutorial guide and a tour of external resources. In our GitHub repository, we offer code examples for all papers whose methods have been implemented in our library. The documentation also features examples of configuration, method setup, model training, and evaluation. 
One such example is visualized in Figure~\ref{fig:code_example}.

\section{Experimental Evaluation}

\textbf{Datasets.}
\method\ evaultes the performance of graph self-supervised methods on six widely-used datasets, including WikiCS~\citep{mernyei2020wiki}, Coauthor~\citep{sinha2015overview}, Amazon-Photo~\citep{mcauley2015image}, IMDB-B~\citep{cai2018simple}, IMDB-M~\citep{zhao2019learning} and Mutag~\citep{niepert2016learning}. The data statistics of these datasets are summarized in Table~\ref{tab:dataset_statistics}.

\begin{table*}[h]
    \centering
    \caption{Dataset Statistics}
    \begin{tabular}{cccccc}
    \toprule 
    Dataset  & \# of graphs & \# of nodes   & \# of edges  & \# of features   & \# of classes      \\ 
    \midrule 
    WikiCS      & 1         &  11,701 & 216,123  & 0 & 10      \\
    Coauthor    & 1         & 18,333 & 163,788 & 6,805 & 15      \\ 
    Amazon      & 1         & 7,650 & 238,162 & 745 & 8      \\ 
    IMDB-B      & 1,000     & 19.8 & 193.1 & 0 & 2      \\
    IMDB-M      & 1,500     & 13 & 26.4 & 0 & 3      \\
    Mutag       & 188       & 17.9 & 39.6 & 7 & 2      \\ 
    \bottomrule
    \end{tabular}
    
    \label{tab:dataset_statistics}
\end{table*}

\begin{table*}[t]
    \centering
    \caption{Reproducibility of graph SSL algorithms on the node classification task.  `-' means that the result is not reported in the research paper.     
    }
\scalebox{0.76}{
    \begin{tabular}{cccccccccc}
    \toprule 
    Methods   &Implementation   & DGI  & GraphCL   & AFGRL & MVGRL & GCA & BGRL & SUGRL & ReGCL    \\ \midrule 
    \multirow{2}{*}{WikiCS}     & Reported     & -       & - & 77.6$\pm$0.5    & -    & 78.3$\pm$0.0 & 80.0$\pm$0.1    & -     & -    \\  
    \cmidrule{2-10}
                                & Ours           & 75.3$\pm$1.1    & 77.3$\pm$0.8 & 78.9$\pm$0.6   & 72.7$\pm$0.9  & 77.6$\pm$0.5 & 78.3$\pm$0.7    & 70.0$\pm$1.1    & 78.7$\pm$0.7   \\ \midrule
    \multirow{2}{*}{Coauthor}    & Reported     & -    & -     & 93.3$\pm$0.2    & -    & 93.1$\pm$0.0 & 93.3$\pm$0.1    & -    & 93.7$\pm$0.3    \\  \cmidrule{2-10}
                                & Ours         & 93.5$\pm$0.3   & 84.0$\pm$0.6    & 92.1$\pm$0.2    & 85.5$\pm$0.4    & 91.6$\pm$0.2 &  92.4$\pm$0.2   & 92.9$\pm$0.1  & 91.2$\pm$0.2  \\ \midrule
    \multirow{2}{*}{Amazon}      & Reported     & -    & -     & 93.2$\pm$0.3    & -    & 92.5$\pm$0.1  & 93.2$\pm$0.3    & 93.2$\pm$0.2  & 92.6$\pm$0.3     \\  \cmidrule{2-10}
                                & Ours         & 92.3$\pm$0.7    & 81.7$\pm$0.9    & 92.6$\pm$0.3  & 86.5$\pm$1.2     & 91.2$\pm$0.3  & 92.6$\pm$0.3     & 91.7$\pm$0.4  & 91.6$\pm$0.4     \\ 
    \bottomrule
    \end{tabular}}
    \label{tab:node_classification}
\end{table*}


\begin{table*}[t]
    \centering
    \caption{Reproducibility of graph SSL algorithms on the graph classification task. `-' denotes that the result is not reported in the research paper. 
    }
    \begin{tabular}{ccccc}
    \toprule 
    Methods   &Implementation   & InfoGraph  & GraphCL   & GraphMAE      \\ \midrule 
    \multirow{2}{*}{IMDB-B}     & Reported     & 73.0$\pm$0.9    &  71.1$\pm$0.4        & -         \\  \cmidrule{2-5}
                                & Ours         &  72.2$\pm$2.1   & 71.5$\pm$3.6      & 72.6$\pm$3.1         \\ \midrule
    \multirow{2}{*}{IMDB-M}    & Reported     & 49.7$\pm$0.5     & -    & -      \\  \cmidrule{2-5}
                                & Ours         & 49.5$\pm$2.7   & 48.0$\pm$3.8        & 49.6$\pm$2.5         \\ \midrule
    \multirow{2}{*}{Mutag}      & Reported     & 89.0$\pm$1.1    &  86.8$\pm$1.3        & -       \\  \cmidrule{2-5}
                                & Ours         & 88.8$\pm$5.0    &  86.7$\pm$4.6       & 86.2$\pm$4.1         \\
    \bottomrule
    \end{tabular}
    \label{tab:graph_classification}
\end{table*}


To demonstrate that the implemented methods can indeed reproduce the performance in the original papers, we evaluate the task performance of all graph self-supervised learning methods implemented in the library. We report the mean and standard deviation for each result, where ±0.0 indicates that the standard deviation is less than 0.05\%. Our implementations generally reproduce the results in the original papers, but slight differences (negligible considering standard deviations) may occur due to data splits or random seeds.  In addition, we provide insights into which methods are preferred for node classification and graph classification tasks.

\textbf{Node Classification.}
Table \ref{tab:node_classification} presents the performance of various graph self-supervised learning methods on the node classification task across three datasets. The following observations can be made: (1) Some methods excel on specific datasets but do not perform consistently well across all datasets. Specifically, SUGRL achieves the highest performance on the Coauthor dataset but performs poorly on the WikiCS dataset, whereas MVGRL shows the best performance on the WikiCS dataset. (2) AFGRL, BGRL, and DGI rank among the top three performers on average across the three datasets. Unlike many contrastive learning-based methods (e.g., GraphCL, GCA), AFGRL and BGRL are non-contrastive methods that do not require negative samples during training. AFGRL utilizes simple augmentations, such as node feature and edge masking, while BGRL is augmentation-free, generating an alternative view of a graph by identifying nodes that share local structural information and global semantics. Similarly, DGI explores local-global consistency by maximizing local-global mutual information.

\textbf{Graph Classification.}
Table \ref{tab:graph_classification} shows the performance of numerous graph self-supervised learning methods on the graph classification task. Notice that though GraphMAE reports the accuracy on these three datasets, GraphMAE only calculates the F1 score in their source code implementation. Thus, we do not include the reported score in Table \ref{tab:graph_classification}. By observation, we find that GraphMAE outperforms InfoGraph and GraphCL on the IMDB-B and IMDB-M datasets. We attribute this superior performance to the design of predicting masked features, which addresses some of the limitations inherent in generative self-supervised graph pretraining methods.

\section{Conclusion}
In this paper, we present the Graph Self-Supervised Learning Toolkit (\method), a user-friendly library designed to streamline and advance research in graph self-supervised learning. Our toolkit offers a comprehensive solution for implementing and experimenting with a wide array of state-of-the-art graph SSL methods.  With robust support for various graph types and a unified framework that simplifies tasks such as dataset loading, hyperparameter tuning, model training, and performance evaluation, \method\ empowers both beginners and experienced practitioners to efficiently explore and apply graph SSL techniques. The evaluations of numerous graph SSL methods included in \method\ offer practical insights for selecting the most suitable approaches for different tasks.  \method\ not only stands as a significant advancement in graph SSL but also complements other related libraries, fostering a collaborative and evolving landscape in graph-based research and applications.

\bibliographystyle{unsrtnat}
\bibliography{sample}

\end{document}